\documentclass[journal]{IEEEtran}
\ifCLASSINFOpdf
\else
\fi

\usepackage[table, dvipsnames]{xcolor}
\usepackage{times}
\usepackage{epsfig}
\usepackage{graphicx}
\usepackage{amsmath}
\usepackage{amssymb}
\usepackage{cite}

\usepackage{pifont}
\newcommand{\cmark}{\text{\ding{51}}}

\usepackage{subfig}
\usepackage{multirow}
\usepackage{makecell}
\usepackage{algorithm,algorithmicx}
\usepackage[noend]{algpseudocode}
\usepackage{soul}
\usepackage{xcolor}

\newcommand*\Let[2]{\State #1 $\gets$ #2}
\algnewcommand\algorithmicoutput{\textbf{Output:}}
\algnewcommand\OUTPUT{\item[\algorithmicoutput]}

\algnewcommand\algorithmicinput{\textbf{Input:}}
\algnewcommand\INPUT{\item[\algorithmicinput]}
\algnewcommand\algorithmicblank{\textbf{     }}
\algnewcommand\BLANK{\item[\algorithmicblank]}
\newcommand{\etal}{\textit{et al.}}
\newcommand{\eg}{\textit{e.g.}}
\newcommand{\ie}{\textit{i.e.}}

\begin{document}
\title{Automatic Video Object Segmentation via Motion-Appearance-Stream Fusion and Instance-aware Segmentation}

\author{Sungkwon~Choo,
        Wonkyo~Seo,
        and~Nam~Ik~Cho,~\IEEEmembership{Senior Member,~IEEE}% <-this % stops a space
\thanks{S. Choo was with the Dept. of Electrical and Computer Eng., and also affiliated with INMC, Seoul National University, Seoul, Korea. W. Seo and N. I. Cho are with the Dept. of Electrical and Computer Eng., and also affiliated with INMC, Seoul National University, Seoul, Korea e-mail: nicho@snu.ac.kr.}% <-this % stops a space
\thanks{Manuscript received MONTH 00, 2019; revised MONTH 00, 2019.}}


\markboth{IEEE TRANSACTIONS ON CIRCUITS AND SYSTEMS FOR VIDEO TECHNOLOGY,~Vol.~00, No.~0, MONTH~2019}%
{Shell \MakeLowercase{\textit{et al.}}: Bare Demo of IEEEtran.cls for IEEE Journals}



\maketitle

\begin{abstract}
This paper presents a method for automatic video object segmentation based on the fusion of motion stream, appearance stream, and instance-aware segmentation. The proposed scheme consists of a two-stream fusion network and an instance segmentation network. The two-stream fusion network again consists of motion and appearance stream networks, which extract long-term temporal and spatial information, respectively. Unlike the existing two-stream fusion methods, the proposed fusion network blends the two streams at the original resolution for obtaining accurate segmentation boundary. We develop a recurrent bidirectional multiscale structure with skip connection for the stream fusion network to extract long-term temporal information. Also, the multiscale structure enables to obtain the original resolution features at the end of the network. As a result of two-stream fusion, we have a pixel-level probabilistic segmentation map, which has higher values at the pixels belonging to the foreground object. By combining the probability of foreground map and objectness score of instance segmentation mask, we finally obtain foreground segmentation results for video sequences without any user intervention, i.e., we achieve successful automatic video segmentation. The proposed structure shows a state-of-the-art performance for automatic video object segmentation task, and also achieves near semi-supervised performance.
\end{abstract}

\begin{IEEEkeywords}
Video object segmentation, Convolutional RNN, Instance-aware segmentation, Two-stream fusion, Spatiotemporal information.
\end{IEEEkeywords}

% \fi
\IEEEpeerreviewmaketitle

\section{Introduction}
\IEEEPARstart{V}ideo object segmentation is to find pixels that belong to objects-of-interest in videos, which requires both temporal and spatial information. Approaches to video object segmentation can be roughly divided into semi-supervised~\cite{bao2018cnn,osvos,Cheng_favos_2018,Li18attention,osvoss,onavos} and unsupervised~\cite{hu2018Mosal,ARP,li2018instance_embed,Li_2018_ECCV_BNN,PDB,Tokmakov17LVO} learning methods. In the field of video object segmentation, the \textit{(un)supervised} means, unlike the general meaning of (not) using the ground truth labels for the training, specific objects are (not) pre-defined by users. More specifically, the semi-supervised method provides a target mask at the first frame, which is then tracked throughout the sequence. The unsupervised method detects foreground objects \textit{automatically} without annotation and continues to detect them in subsequent frames. Hence, the semi-supervised methods show relatively higher performances, while the unsupervised methods have the advantage that they do not need cumbersome user intervention.

In this paper, we develop a new unsupervised video object segmentation algorithm based on the instance segmentation and \textit{two-streams fusion} scheme~\cite{cheng2017segflow,hu2017maskrnn,jain2017fusionseg,Tokmakov17LVO}, which use motion and appearance information from two separate networks.
One of the main ideas of our method is to extract long-term temporal information and multiscale spatial information as the motion and appearance stream, respectively, for accurate foreground segmentation.
For this, we construct the stream networks to have recurrent encoder-decoder structures, which learn temporal and spatial weights of various scales unlike the conventional shallow or fixed-shape convolutional recurrent networks~\cite{luo2017unsupervised,qiu17dstfcn,tseng2017joint,ond16Deeptracking_aaai,convlstm}. Also, we make the network satisfy both bidirectional and cascaded characteristics for maintaining spatial and sequential information as much as possible.
While the conventional approaches fuse the motion and appearance with the reduced resolution (in the middle of the networks), the proposed structure fuses the two-streams after each stream is restored to the original resolution (at the end of proposed multiscale recurrent network). As s result, we obtain the pixel-level probabilistic foreground map. Finally, we integrate the foreground map with the objectness score of instance segmentation mask for obtaining the video object segmentation results.

Although it seems reasonable to use instance information for the video segmentation, it is rarely used in unsupervised methods~\cite{Seguin16,xiao2016track}.
This is because, unlike the semi-supervised methods, there are no annotations that can exclude false positives when there are multiple instances. Since the unsupervised method is not requiring intervention to annotate the object of interest or to eliminate the false positives, we calculate the foreground probability of the instance using the pixel-level segmentation results from the stream fusion and objectness scores of masks. By boosting the instance segmentation area with the highest pixel-level probability and score, the main objects in the video sequence are revealed.
From the experiments on popular datasets, it is shown that our network outperforms the existing unsupervised methods. Moreover, it shows comparable performance to the state-of-the-art semi-supervised methods.

\section{Related work}

\subsection{Recurrent network in video segmentation}
\label{related:recurrent}
The recurrent network was proposed to discover temporal relations of consecutive frames in videos~\cite{milan2017online,Pigou2018,zhou2017see}, which finds many applications, including video object segmentation. For example, Tokmakov~\etal~\cite{Tokmakov17LVO} delivered both optical flow information and visual features to the convolutional Gated Recurrent Unit (ConvGRU)~\cite{cho2014learning} to obtain segmentation results. Song~\etal~\cite{PDB} predicted the object region by stacking the bidirectional LSTM for the visual features. Li~\etal~\cite{Li18attention} predicted the mask in the next frame through the recurrent network for each object. In these methods, they adopted relatively shallow recurrent networks such as a pre-trained ResNet~\cite{resnet} or VGG~\cite{Simonyan14VGGNet}. Hence, these methods have some drawbacks in that the shallow network does not provide a large receptive field and long-term temporal information. Also, features from the conventional VGG or ResNet have reduced resolution, and hence do not possess the details of shapes.
For the long-term temporal prediction, it was validated in ~\cite{PascanuGCB13} that the recurrent network at each level has a different timescale as the number of stacks grows. Also, some studies~\cite{finn2016unsupervised,lotter2016deep} demonstrated that a deep stack of recurrent networks could provide a long-term prediction. But, these structures were used only for the unsupervised video prediction.
To overcome the above-stated limitations to be used for the video segmentation, we propose a multiscale recurrent network which has various levels of the recurrent network and the encoder-decoder structure. Specifically, we stack multiple levels of recurrent networks in a cascaded bidirectional manner to provide a large time scale. Also, the encoder-decoder structure minimizes the loss of spatial information by taking the optical flow and images as inputs and produces the spatiotemporal information of the original resolution.

\subsection{Two-stream fusion}
In general, training a deep network for video applications is not easy because the annotated data is fewer than the case of still-image. 
Also, the information from adjacent frames is usually redundant, {\em i.e.,} there are relatively few distinct frames from even a long video sequence. To cope with this problem, we exploit a pretrained network such as ImageNet~\cite{IMAGENET} or COCO~\cite{lin2014microsoftcoco}. But the video applications need to use not only the spatial contents but also the temporal information which may be more critical at times. Hence, there have also been many studies to use temporal information. Especially, there are {\em two-stream fusion} approaches that combine information from two separate networks where one extracts temporal (motion) and the other spatial (appearance) features from a video sequence~\cite{karpathy2014large,simonyan2014two,feichtenhofer,gao,ouyang,kwon,gammulle,zeng}. These methods were developed mostly for video classification, such as action recognition and person re-identification. For example,
Karpathy~\etal~\cite{karpathy2014large} proposed a two-stream structure for video under various conditions, which combines the information of two consecutive frames at the end of separate networks. They showed that this {\em late} fusion provides better performance than the {\em early} fusion that combines the consecutive frames before the end of the network. Simonyan and Zisserman~\cite{simonyan2014two} further improved the performance of two-stream fusion by feeding the optical flow to a network and the input image to the other. Hence, the first network provides the stream of temporal information and the other the appearance stream.

Most of the recent studies~\cite{carreira2017quo,hu2017maskrnn,tran2015learning,wang2016temporal} showed better performance when two streams were used independently using the late fusion. However, the late fusion causes loss of resolution due to the reduced scale. Hence, a multiscale refinement of two-stream fusion~\cite{hu2018fully} was also proposed to restore full resolution.
Our experiment also confirms that the late fusion performs well in a two-stream structure without interfering with the learning of each other's stream.
In summary, the proposed encoder-decoder fusion is designed to combine the streams after restoring the size to the original resolution, thereby reducing information loss and obtaining more accurate results.

\subsection{Instance-level video segmentation}

In the case of semi-supervised segmentation where the annotated ground truth is given in the start frame, it is reasonable to use the instance-level object segmentation for the rest of the frames, with the initial annotation or results of the previous frames as instance proposals. Specifically, we may employ MaskRNN~\cite{hu2017maskrnn} for object region detection and location tracking for each frame in a video. For some examples of video segmentation that explicitly use the instance segmentation or region proposal, OSVOS-S~\cite{osvoss} use a region proposal network to find multiple instance proposals and select those that match the ground truth of the first frame. Continuing the following frames, OSVOS~\cite{osvos} method selects the next proposal and detects the object.  Li~\etal~\cite{Li18attention} obtain object regions and features by re-identification (Re-ID) with the ground truth as the starting point, and find the next region of the object in the recurrent network. FAVOS~\cite{Cheng_favos_2018} utilizes a part-based tracking method and aggregates the tracked parts to generate the final segmentation. This method performs well without online learning and shows the fastest segmentation speed with 0.6 second execution time.

In the case of unsupervised methods, there are few studies that detect video objects at instance-level. Xiao~\etal~\cite{xiao2016track} detect object proposals with high objectness, then cluster these proposals spatiotemporally and iteratively grow the clusters. Their results did not surpass all supervised methods but showed better results than many baselines without human annotation. In this paper, we propose an instance-aware segmentation method that finds the main object of video with pixel-level segmentation and uses it to select an instance proposal. That is, the proposed method can segment the main foreground objects without mask annotation.

\begin{figure*}[h!]
\centering
\includegraphics[width=.99\linewidth]{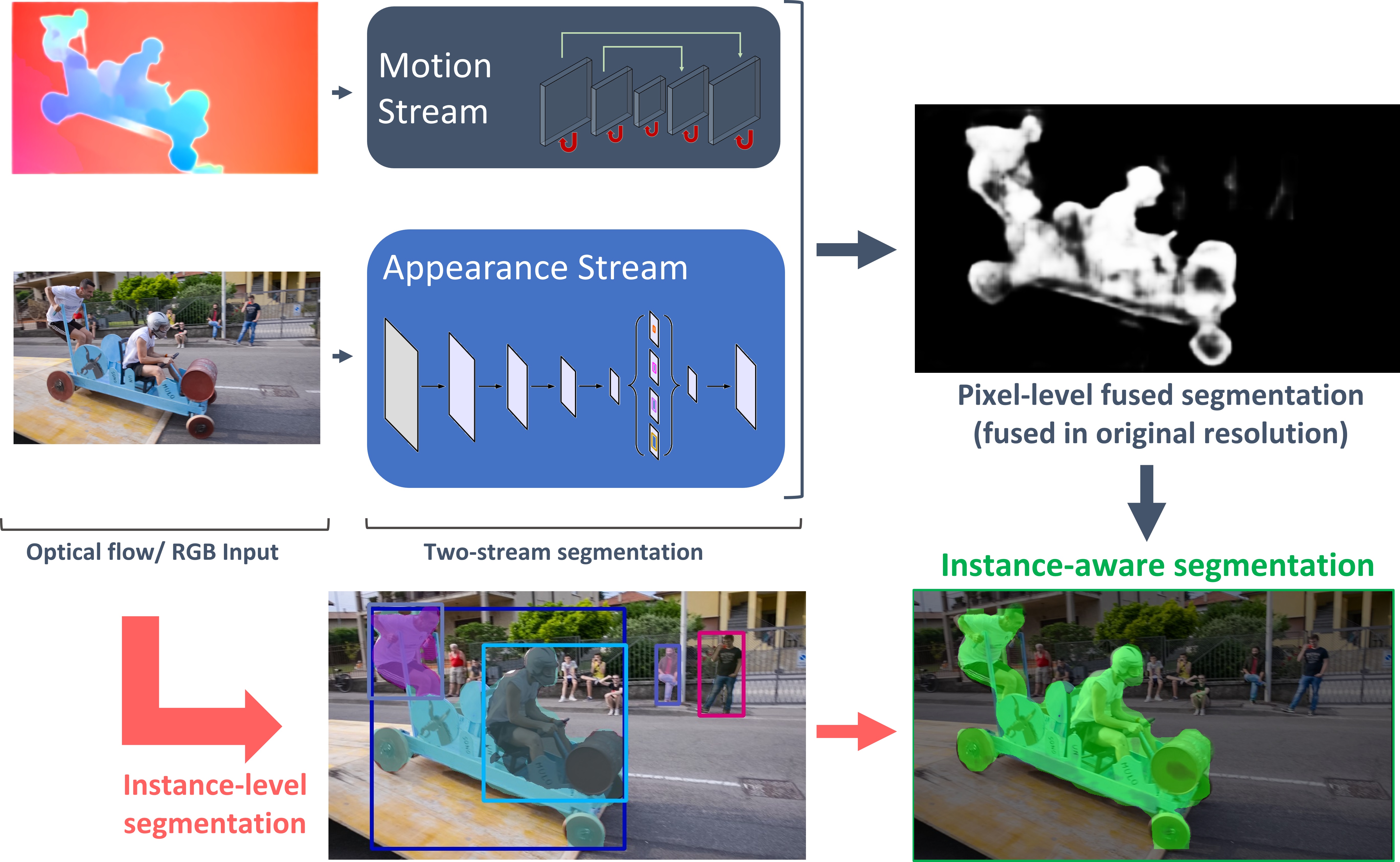}
\caption{The overview of the proposed network.
The two networks in the left-upper side extract motion stream and appearance stream from the video respectively, which consist of encoder-decoder network. The motion stream network receives the optical flow, and the appearance network gets the original RGB input frame by frame. The two streams are merged at the end of the decoder to keep the resolution, which results in the pixel-wise probability of belonging to the foreground moving object (pixel-wise foreground map in the right-upper part). The left-lower part shows the instance segmentation network that finds several instance-level segments in the scene. The instance-level segments are ranked according to the two-stream fusion map, which gives the final foreground segment, as shown in the right-lower picture.}
\label{fig:archi}

\end{figure*}

\section{Method}

The modules and flow of our method are described in Fig.~\ref{fig:archi}. We will explain the proposed approach by explaining the role of modules and the combination of their outputs.

\subsection{Motion Stream Network}
\label{sec:msgru}
The motion stream network extracts features related to temporal information for each frame, where the sequence of these features for the consecutive frames is the {\em motion stream}. It has the encoder-decoder structure as shown in the figure, which takes as input the optical flow calculated by FlowNet 2.0~\cite{ilg2017flownet}. In the existing video object segmentation studies, motion stream was acquired either through handcrafted design (\eg~temporal clustering~\cite{xiao2016track}, flow warping~\cite{li2017video}) or through a shallow recurrent network~\cite{PDB,Tokmakov17LVO}. Compared to the existing models, we use the multiscale bidirectional architecture to learn the sufficient sequential information of motion sequences. In the following, we explain the elements that constitute the motion stream network.

\subsubsection{Multiscale ConvGRU} 
We design the motion stream network as an hourglass-shaped recurrent network with skip connection. This structure has the advantage of securing various timescales with minimal loss of resolution while ensuring sufficiently large spatial receptive fields.
To be more precise, we construct it by stacking the $L$ convolutional GRUs (ConvGRU). The stride of convolution is set to 2 so that the scale becomes smaller at each layer of the encoder side, and features are resized by bilinear interpolation when the scale is increased at the decoder side. In summary, Algorithm~\ref{code:archi} shows the signal flow in our multiscale structure, where $[ \cdot \, ; \cdot ]$ means the concatenation on the channel axis, $T$ is the sequence length of the mini-batch, and $X_t$ is the input to the multiscale GRUs which differ depending on the direction of GRU as will be explained later.

\setlength{\textfloatsep}{10pt}% Remove \textfloatsep
\begin{algorithm}
  \caption{Multiscale cascaded bidirectional GRU.}
  \begin{algorithmic}[1]
    \INPUT{Optical flow $F =\{F_{t+1},...,F_{t+T} \}$,}
    \BLANK{\hspace{16pt}Previous GRU states $H_{t} = \{h_{t}^1,...,h_{t}^L\}$}
    \OUTPUT{Motion stream feature $z = \{z_{t+1},...,z_{t+T} \}$,}
    \BLANK{\hspace{22pt}Updated states $H_{t+T} = \{h_{t+T}^1,...,h_{t+T}^L\}$}
    \Statex
    % \Function{MultiScaleGRU}{$F_t, H_{t-1}$}
    \Function{MultiScaleGRU}{$X_t, H_{t-1}$}
      \Let{$x_t^1$}{$X_t$} 
      \For{$l \gets 1 \textrm{ to } (L+1)/2$} %\Comment{nEncLayer : number of encoder layers}
          \Let{$h_{t}^{l}$}{ConvGRU$^l$($x_t^{l},h_{t-1}^{l}$)} \Comment{stride=2}
          \Let{$x_{t}^{l+1}$}{$h_{t}^{l}$} 
      \EndFor

      \For{$l \gets (L+1)/2+1 \textrm{ to } L$} %\Comment{nEncLayer : number of decoder layers}
          \Let{$x_{t}^{l}$}{resizeBilinear($x_{t}^{l}$)} \Comment{ratio=2}
          \Let{$h_{t}^{l}$}{ConvGRU$^l$($x_t^{l},h_{t-1}^{l}$)} 
          \Let{$x_{t}^{l+1}$}{$[h_{t}^{l} ; h_{t}^{L-l+1}]$}  \Comment{skip connection}
      \EndFor

      \Let{$o_{t}$}{transposedConvolution($x_{t}^{L+1}$)} \Comment{stride=2}
      \State \Return{$o_{t}$, $H_t$}
    \EndFunction
    \Statex

    \Function{CascadedBidirectionalGRU}{$F, H_{t}$}

      \For{$s \gets 1 \textrm{ to } T$}
          \Let{$o_{s}^{F},H_{t+s}^F$}{MultiScaleGRU$^F$($F_{t+s},H_{t+s-1}$)} %\Comment{forward}
      \EndFor

      \Let{$I^{B}$}{reverse($[o_{1,..,T}^F;F]$)} 
      \Let{$H_{0}^{B}$}{$H_{t+T}^{F}$} 

      \For{$s \gets 1 \textrm{ to } T$} 
          \Let{$o_{s}^{B},H_{s}^B$}{MultiScaleGRU$^B$($I_s^B,H_{s-1}^B$)} 
      \EndFor

      \Let{$z$}{$[o_{1,..,T}^F;$ reverse($o_{1,..,T}^B$) $]$} 

      \State \Return{$z$, $H_{t+T}$}
    \EndFunction
  \end{algorithmic}
\label{code:archi}
\end{algorithm}

In the implementation, we use the asymmetric convolution to increase the receptive field with less computations, {\em i.e.,} all ConvGRUs are designed as $1 \times k$ and $ k\times1$ convolutions. The features of both combinations are calculated and stacked to eliminate the influence of calculation order, and the weights of two combinations are learned individually. Then, the layer normalization~\cite{ba2016layernorm} is applied separately for input and state of ConvGRU, which is expressed as 
\begin{align}
  &\begin{aligned}
  &g_{s} = LN\left(\begin{bmatrix}W^{g,0}_{s,k \times 1} *W^{g,0}_{s,1 \times k} * s_t;\\
  W^{g,1}_{s,1 \times k} *W^{g,1}_{s,k \times 1}* s_t \end{bmatrix}\right)\\
  & \text{where } s \in\{ \text{input } x,\text{state } h \},
  \end{aligned} \\
  &\begin{aligned}
  g = \sigma(g_x + g_h)
  \end{aligned} 
\end{align}
where $LN(\cdot)$ is for the layer normalization, $g$ is for the reset gate $r$ and update gate $z$, * is the convolution operation, and $\sigma(\cdot)$ is for sigmoid function. This structure is chosen experimentally because the superiority of the various implementations of the input and state combinations and the position of the normalization layer was not proven.
Also, state candidate in our model is calculated by the same asymmetric operation with the gates and then updated to generate the output, which is described as
\begin{align}
  &\begin{aligned}
  &c_{s} = LN\left(\begin{bmatrix}W^{c,0}_{s,k \times 1} *W^{c,0}_{s,1 \times k} * s_t;\\
  W^{c,1}_{s,1 \times k} *W^{c,1}_{s,k \times 1}* s_t\end{bmatrix}\right) \\
  & \text{where } s \in\{ \text{input } x,\text{reset state } q = r \odot h_{t-1} \},
  \end{aligned}\\
  &\begin{aligned}
  &\tilde{h} = \mathrm{tanh}(c_x + c_q), \\
  &h_t =z \odot h_{t-1} + (1-z) \odot \tilde{h}
  \end{aligned} 
\end{align}
where $\odot$ denotes pixel-wise multiplication operation (the Hadamard product) and $\mathrm{tanh}(\cdot)$ is for hyperbolic tangent function. The effectiveness of this multiscale recurrent network design can be seen in Table~\ref{table:Ablation_RNN}, which will be explained in detail in the experiment section. The results mean that although the optical flow itself provides inter-frame information, longer temporal dependency information from the stream network helps to obtain much better performance.
For additional evaluation, performance according to the number of GRUs is shown in Table~\ref{table:Ablation_more}.

{\renewcommand{\arraystretch}{1.2} 
\begin{table}

\caption{Variations of motion stream architecture and its performance change. Bi-GRU stands for bidirectional GRU and Cas-Bi-GRU stands for cascaded bidirectional GRU. All networks in the table consist of only the motion stream. As the timescale of the structure increases, the performance improves.}
\begin{center}
\begin{tabular}{|l|c|c|c|}
\hline
Variant &  $\mathcal{J}$ Mean  & $\mathcal{F}$ Mean & $\mathcal{J}\&\mathcal{F}$ Mean    \\
\hline
2D Conv&64.98 & 63.38  &  64.18  \\
 3D Conv  &67.32 &66.48 & 66.90   \\
 GRU & 67.61 & 68.15   &  67.88  \\
 Bi-GRU & 69.32&69.89   & 69.60   \\ 
 Cas-Bi-GRU & 70.05 &69.70   & 69.88   \\ 
\hline
\end{tabular}
\end{center}

\label{table:Ablation_RNN}
\end{table}
}

{\renewcommand{\arraystretch}{1.2} 
\begin{table}
\caption{Results according to RNN stacks. The figures of merit are in $\mathcal{J}\&\mathcal{F}$.}
% \vspace{-19pt}
\begin{center}
\setlength\tabcolsep{5 pt}
\begin{tabular}{|l|c|c|c|c|c|}
\hline
% \# of GRU& 0 & 1 & 3 (ss) & 3 & 5    \\
% \hline 
% Motion stream& 64.18 &66.07 &  66.94 & 68.27 & 68.28  \\
% M + A stream& 72.04 &74.46 &  74.51& 74.72 & 75.22  \\
\# of GRU& 0 & 1 & 3 & 5    \\
\hline 
Motion stream& 64.18 &66.07  & 68.27 & 68.28  \\
M + A stream& 72.04 &74.46 & 74.72 & 75.22  \\
\hline
\end{tabular}
\end{center}
 % As the timescale of the structure increases, the performance improves}
\label{table:Ablation_more}
\end{table}
}

\subsubsection{Cascaded Bidirectional Network}  
In addition to being a multiscale structure as explained above, our network is designed to work bidirectionally for increasing the performance through the longer temporal connections, as described in Algorithm~\ref{code:archi}. This structure is similarly proposed in PDB~\cite{PDB}, where the cascade network forwards only the output of the previous network to the input of the backward network. Unlike this method, we stack the optical flow input and the output of the forward network for the backward network.
% :  
% \begin{equation}
% \begin{aligned}
% &x_{t}^{backward, 0} = [x_{T-t+1}^{0}; h_{T-t+1}^{L}]
% \end{aligned}
% \end{equation}
% where $T$ is the sequence length of the mini-batch. 
Also, the proposed architecture is designed to share the state information among cascade networks, while state information is continuously transmitted. 
%It consists of two state transfers. 
The last state of the forward GRU is transferred to the next batch so that the state of the first frame can be transmitted until the last frame of the video. The backward GRU also uses the state of the forward GRU as the initial state.
% These transfers are expressed as
% \begin{equation}
% \begin{aligned}
% h_{0}^{f} &= h_{T}^{f,\text{previous mini-batch}} &  \\ 
% h_{0}^{b} &= h_{T}^{f} &\text{for } l>0~. 
% \end{aligned}
% \end{equation}
This transfer scheme, called \textit{stateful}, is efficient if the length is variable or longer than the memory limit. It minimizes the problem caused by the state initialization of the bidirectional GRU and allows continuous state information of the video to be used in the backward network. It also shows good performance compared to the \textit{stateless} result and generates seamless results between batches without a redundant operation of the sliding window method.

\subsection{Appearance Stream Network}
\label{sec:fusion}
In addition to the motion stream, we extract a stream of appearance information from the video.
We adopt the structure of DeepLabv3+~\cite{deeplabv3plus2018} as our appearance stream network, which is one of the state-of-the-art semantic segmentation architectures. This is also in the shape of encoder-decoder design to make predictions close to the original resolution. After initializing the network with the learned weights of existing semantic segmentation task (PASCAL VOC 2012~\cite{pascal-voc-2012}), we further train it to learn the target dataset. However, since datasets for video object segmentation consist of relatively few video and redundant frames, training the entire network with this dataset results in poor validation performance due to overfitting. To prevent this, we update the weight only for the decoder.
% , not the entire structure. 

\begin{figure}

\centering
\setlength\tabcolsep{1.5 pt}
\begin{tabular}{lccccc}

\includegraphics[width=.49\linewidth]{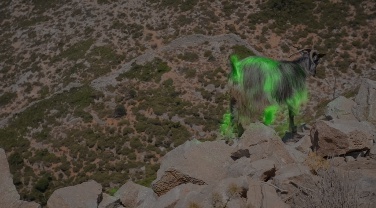}&
\includegraphics[width=.49\linewidth]{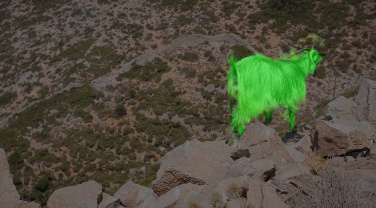}\\
\includegraphics[width=.49\linewidth]{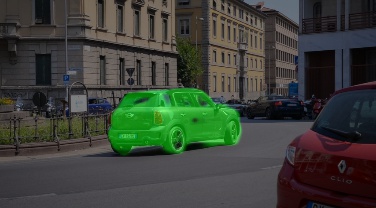}&
\includegraphics[width=.49\linewidth]{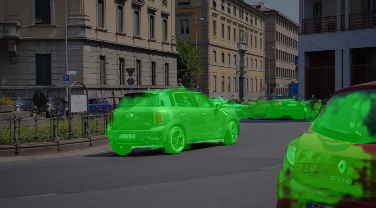}\\
\end{tabular}

\caption{An example that shows the activation characteristics of each stream (\textit{left}: motion stream, \textit{right}: appearance stream). In \textit{goat} sequence, the motion stream fails because the object has a small motion relative to the camera motion. In the \textit{car-roundabout} sequence, the motion stream detects only the moving object among several cars, but the appearance stream detects all the moving and parked cars.}
\label{fig:rnnvssem}
\end{figure}

\subsection{Fusion of Two Streams in the Original Resolution}
Since the two streams are trained by the motion and appearance information, respectively, the results show different characteristics.
As can be seen in Fig.~\ref{fig:rnnvssem}, the motion stream network detects the object with larger motion than the others, {\em i.e.,} it finds the region with a higher probability of being a foreground object. However, the network fails to detect the main object when it has small or similar motion compared to the background or surrounding objects.
On the other hand, the appearance stream network detects areas with high objectness regardless of motion information, and thus highlights many unimportant objects that belong to the background.
% We have devised various methods for fusion of two-stream.
% In the two-stream fusion studies, average, SVM or network fusion was proposed~\cite{simonyan2014two}. 

% In the video object segmentation studies, a late fusion method is used to maintain the spatial dimension in order to obtain the segmentation result~\cite{Tokmakov17LVO,PDB}. 
To complement each other's weakness, we combine the two streams.
We fuse two streams after each network reconstructs features to the original size through the decoder, unlike the previous methods that concatenate features at the reduced resolution and then feed them to more convolution layers.
Specifically, the proposed method concatenates the streams in the channel axis and then combine them through a 1x1 convolution.
Table~\ref{table:Ablation_Resolution} compares the performance of these two schemes, where we can see that the proposed method reduces the loss of information compared to the previous two-stream fusion methods.

{\renewcommand{\arraystretch}{1.2} 
\begin{table}

\caption{Comparison of stream fusion results. The conventional fusion with reduced resolution can be considered a fusion at the encoder, which is shown to provide lower precision $\mathcal{F}$ than the proposed method (both before/after the CRF).}
\begin{center}
\begin{tabular}{|lc|c|c|c|c|}
\hline

\multicolumn{2}{|c|}{Variant} & Encoder & Enc+CRF & Ours & Ours+CRF \\
\hline 
\hline
% $\mathcal{J}$ 
% & Mean $\mathcal{M}  \uparrow$    
\multicolumn{2}{|c|}{$\mathcal{J}$ Mean $\mathcal{M}  \uparrow$}
 &75.32     &78.89    & 76.86    & 79.73\\
% $\mathcal{J}$ & Recall $\mathcal{O} \uparrow$     &90.09     &92.57    & 90.73    & 93.23\\
% & Decay $\mathcal{D} \downarrow$                  &5.33     &5.43    & 5.04    & 4.27 \\
\hline   

% $\mathcal{F}$                                         
% & Mean $\mathcal{M} \uparrow$                     
\multicolumn{2}{|c|}{$\mathcal{F}$ Mean $\mathcal{M}  \uparrow$}
&74.55     &76.22    & 77.27    & 77.79\\
% $\mathcal{F}$ & Recall $\mathcal{O} \uparrow$     &85.54     &86.31    & 89.83    & 87.81\\
% & Decay $\mathcal{D} \downarrow$                  &5.70     &5.44    & 5.27    & 3.41 \\
\hline                                            
\multicolumn{2}{|c|}{$\mathcal{J}\&\mathcal{F}$ Mean }  &74.93 &77.55 &77.06 &78.76  \\
\hline
\end{tabular}
\end{center}

\label{table:Ablation_Resolution}
\end{table}
}
\subsection{Instance-aware segmentation}
\label{section:instance}
We use Mask R-CNN~\cite{he2017mask} for the instance detection, where we add a feature pyramid network~\cite{lin2017feature} and a cascade structure. Based on the pretrained weights for the COCO dataset~\cite{lin2014microsoftcoco}, additional training is conducted using the video object segmentation dataset. Although the learned network for COCO shows a fairly high detection rate, it needs to be fine-tuned because it fails to detect objects that are not in the class vocabulary. Additional training is designed to learn the classification, bounding box regression, and mask segmentation for a single foreground class, rather than the 80 classes of COCO.
The fine-tuned instance proposal network achieves a significant performance by detecting only a single instance with the highest objectness score (the third column of Table~\ref{table:Ablation} that will be explained in the ablation study). However, this result is the largest or most obvious object in the current frame, regardless of the temporal context. Also, since the segmentation result of Mask R-CNN is extracted at a very low resolution, direct use of this result reduces the segmentation performance significantly.
Hence, we devise following methods for the effective integration of two segmentations.

1. Calculate the IoU (Intersection over Union) of each detected instance $M^i_{obj}$ (a binary image where $M^i_{obj}(x)=1$ if the pixel $x$ belongs to the $i$-th mask, and 0 elsewhere) for $i \in\{1,...,N_{obj}\}$ and pixel-level segmentation result
$M_{pixel}$ (real value between 0 and 1):
\begin{equation}
IoU^i_{obj} = \frac{\sum_x  M^i_{obj}(x) \cdot M_{pixel}(x)}{\sum_x \max(M^i_{obj}(x), Bin(M_{pixel}(x)))}~
\end{equation}
where the summation is over the image, and $ Bin (\cdot) $ is the binarization with threshold 0.5. 

2. Define scores to each instance:
\begin{equation}
Score^i = IoU^i_{obj} \cdot objectness^i~
\label{eq:instance}
\end{equation}
where the $objectness^i$ is the objectness score of the proposal $i$ which is the result of Mask R-CNN.

\begin{figure}
\centering
\setlength\tabcolsep{1.5 pt}
\begin{tabular}{lccccc}
\includegraphics[width=.49\linewidth]{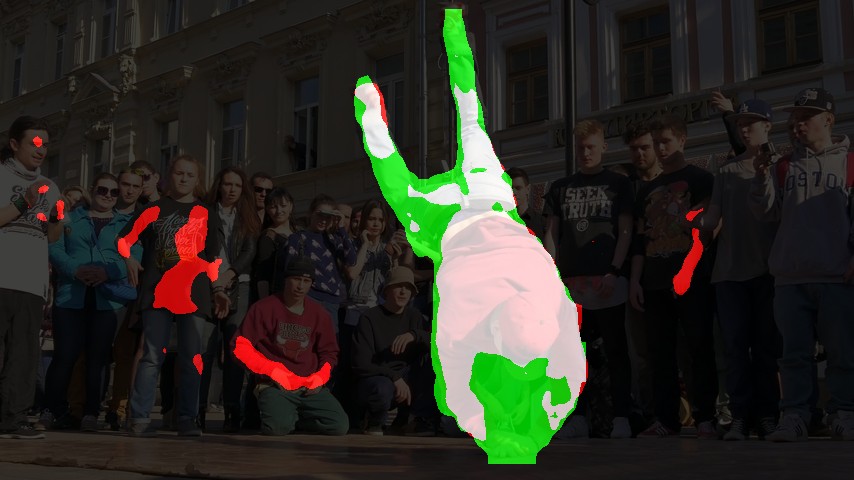}&
\includegraphics[width=.49\linewidth]{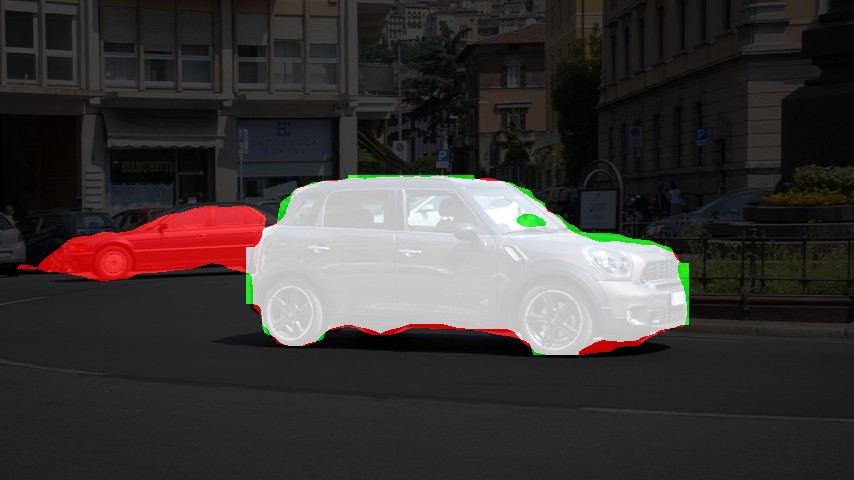}\\
\end{tabular}

\caption{Examples of the instance-aware segmentation. \textit{Red} and \textit{green} are regions detected by pixel-level and instance-level segmentation, respectively. White region is the intersection of red and green regions. With the instance-aware segmentation, we can detect the missed area of the target object (\textit{left}) and exclude other objects (\textit{right}).}
\label{fig:instance}
\end{figure}

3. Boost the region of the instance $p$ with the highest score by the pixel-level segmentation mean within the region. Except for the selected instance area, reduce the pixel-level result by half:
\begin{equation}
\begin{aligned}
&M_{pixel}(x) = 
\begin{cases}
\min(M_{pixel}(x) + \mu_{obj},1) ,& \text{if } x \in M^p_{obj} \\
M_{pixel}(x)/2,     & \text{otherwise}
\end{cases}
% \\&\text{where }  \mu_{obj} =  \left( \sum_{x\in M^p_{obj}}{M_{pixel}(x)} \right)
% /\left( \sum_{x\in M^p_{obj}}{M^p_{obj}(x)} \right)~,
\\&\text{where }  \mu_{obj} =  \frac{\sum_{x\in M^p_{obj}}{M_{pixel}(x)}}{\sum_{x\in M^p_{obj}}{M^p_{obj}(x)}}~,
\end{aligned}
\end{equation}
which removes false positives when there are multiple instances or surrounding clutter in the pixel-level result as in Fig.~\ref{fig:instance}. On the contrary, it serves to compensate for the false negative even for non-rigid objects not detected in the pixel-level results.

{\renewcommand{\arraystretch}{1.2} 
\begin{table*}[!t]
\caption{Evaluation on the validation set of DAVIS dataset~\cite{davis2016} and comparison with state-of-the-art results of semi-supervised and unsupervised methods. The best result for each experimental method is shown in bold. Our results show large performance gaps to the previously proposed unsupervised methods.}
\begin{center}
\begin{tabular}{|lc|c|c|c|c|c|c|c|c|c|c|c|}
\hline
& & \multicolumn{3}{c|}{Semi-supervised}   &  \multicolumn{6}{c|}{Unsupervised}  \\
\hline
&       &  \ OSVOS~\cite{osvos} &     \ OSVOS-S~\cite{osvoss} &   \ OnAVOS~\cite{onavos} &   \ LVO~\cite{Tokmakov17LVO} & PDB~\cite{PDB} & MoSal~\cite{hu2018Mosal} & Embed~\cite{li2018instance_embed}  &BNN~\cite{Li_2018_ECCV_BNN} &\bf\ Ours \\
\hline\hline
% \multicolumn{2}{|c|}{$\mathcal{J}\&\mathcal{F} $ Mean $\uparrow$}   & \ 80.2& \bf\ 86.5& \ 85.5& \ 74.0& \ 75.9& \ 79.4& \bf\ 84.0 \\
% \hline
& Mean $\mathcal{M}                 \uparrow$   &   \ 79.8 &   \ 85.6 &\bf\ 86.1 &   \ 75.9  &   \ 77.2 &   \ 77.6 &   \ 78.5  & \ 80.4 &   \bf\ 84.9 \\
$\mathcal{J}$ & Recall $\mathcal{O} \uparrow$   &   \ 93.6 &\bf\ 96.8 &   \ 96.1 &   \ 89.1  &   \ 90.1&   \ 88.6&   \ - & \ 93.2&   \bf\ 96.8 \\
& Decay $\mathcal{D} \downarrow$                &   \ 14.9 &   \ 5.5 &   \ \bf5.2 &   \bf\ 0.0  &   \ 0.9&   \ 4.4&   \ - & \ 4.8&   \ 1.9 \\
\hline
& Mean $\mathcal{M} \uparrow$                   &   \ 80.6 &\bf\ 87.5 &   \ 84.9 &   \ 72.1  &   \ 74.5&   \ 75.0&   \ 75.5 & 78.5&   \bf\ 83.1 \\
$\mathcal{F}$ & Recall $\mathcal{O} \uparrow$   &\ 92.6 &\bf\ 95.9 &   \ 89.7 &   \ 83.4  &   \ 84.4&   \ 86.9&   \ - & \ 88.6&   \bf\ 89.6 \\
& Decay $\mathcal{D} \downarrow$                &   \ 15.0 &   \ 8.2 &   \ \bf5.8 &   \ 1.3  & \bf\ -0.2& \ 4.2& \ - & \ 4.4&   \ 2.3 \\
\hline
$\mathcal{T}$ & Mean $\mathcal{M} \downarrow$   &   \ 37.6 &   \ 21.4 &   \bf\ 18.5 &   \ 26.5  &   \ 29.1 & \ 24.3&\ -&\ 27.8&   \bf\ 18.7 \\
\hline
\end{tabular}
\end{center}
\label{table:comparison}
\end{table*}
}

\subsection{Implementation details}
\subsubsection{Preprocessing}
As a preprocessing, we match the magnitude of the image and optical flow. Specifically, we normalize the RGB color information to the range of $ [- 1,1] $, and also normalize the optical flow such that its maximum magnitude is 1 within a frame. The mini-batch consists of $ T = 20 $ sequential frames which is resized from $ 854 \times 480 $ to $ 376 \times 208$. Reducing the resolution is a disadvantage, but it is also important that the backpropagation of the recurrent network be learned at a sufficient sequence length. Also, we believe that our reduced resolution is still large compared to the stridden output resolution of the other methods (\eg~$ 1/8 $). During the training, the input is randomly cropped to $ 800 \times 448 $ for data augmentation and then scaled.

\subsubsection{Network}
In the motion stream, a stack of $L=5$ ConvGRUs is used in each direction.
The layer has a stride of $ [2,4,8,4,2] $ relative to the original resolution,
the number of hidden units of the layer is $ [16,64,128,64,16] $, and asymmetric kernel size $ k = 7 $. The output of the appearance stream with DeepLabv3+ structure is $ 1/4 $ the size of the input. We interpolate this result bilinearly without additional convolution operation and transfer it to the stream fusion phase. Also, the optical flow is transmitted not only to the input of the motion stream but also to the stream fusion phase so that the motion stream becomes a very large residual network. The two-stream structure is trained at once using Adam optimization~\cite{kingma2014adam}. Learning rate starts from $ 1e-4 $ and gradually decreases, and we use the gradient clipping to limit the norm not to exceed 5. The network for the instance proposal is learned separately from the two-stream structure. We train a single-class classification based on the Mask R-CNN~\cite{he2017mask} structure and pretrain the weights implemented in the Tensorpack library~\cite{wu2016tensorpack}.
To detect the objects with higher accuracy, we train with the anchor's positive and negative thresholds increased from the original setting of $ (0.7,0.3) $ to $ (0.9,0.5) $.

\subsubsection{Postprocessing}
Most studies use the graph, superpixel, or conditional random field (CRF) for video segmentation results that fit the boundaries of the image. We binarize the results and refine the boundary using the dense CRF~\cite{krahenbuhl2011efficient}. The weight of the bilateral and Gaussian kernel are $ w_1 = 5, w_2 = 3$ and the parameters of CRF are set as $ (\theta_\alpha,\theta_\beta,\theta_\gamma) = (30,5,3)$. In addition, the Kalman filter is used to stabilize the instance proposal results. The choice of the instance proposal is determined by the IoU and instance detection scores of the pixel-level segmentation results such as Eq.~\ref{eq:instance}. However, if the pixel-level segmentation fails to suppress the activation for multiple objects or the score of the instance proposal is incorrectly measured, an object at the wrong position may be detected. In order to prevent this, we limit the instance proposal to only the range predicted by the tracker with the Kalman filter of the linear constant velocity model, as used in~\cite{Bewley2016_sort}. This post-processing yields a stable result yet very efficient. The performance change can be seen in Table~\ref{table:Ablation}.

{\renewcommand{\arraystretch}{1.4} 
\begin{table*}
\caption{Comparison of results for Freiburg-Berkeley Motion Segmentation Dataset (FBMS-59)~\cite{14FBMS}. All the results of this table are taken from the paper, and `-' indicates the results are not shown in the paper. }
\begin{center}
\begin{tabular}{|c|c|c|c|c|c|c|c|c|c|c|c|c|c|c|}

\hline
 Method  & ARP~\cite{ARP}  & LVO~\cite{Tokmakov17LVO} & \ MoSal~\cite{hu2018Mosal} & Embed~\cite{li2018instance_embed} & BNN~\cite{Li_2018_ECCV_BNN} & PDB~\cite{PDB}  &\bf\ Ours \\
\hline
$\mathcal{J}$ Mean & 59.8  & 65.1 & 60.8& 71.9  & 73.9 & 74.0 & \bf 78.3\\
\hline
F-score     & - & 77.8 & - & 82.8  & 83.2 & - &\bf  85.1\\
\hline
\end{tabular}
\end{center}
\label{table:FBMS}
\end{table*}
}

\begin{figure*}[t]
\centering
\setlength\tabcolsep{1.5 pt}
\begin{tabular}{lccccc}

\raisebox{.7\height}{\rotatebox[origin=t]{90}{Breakdance}} &
\includegraphics[width=.19\linewidth]{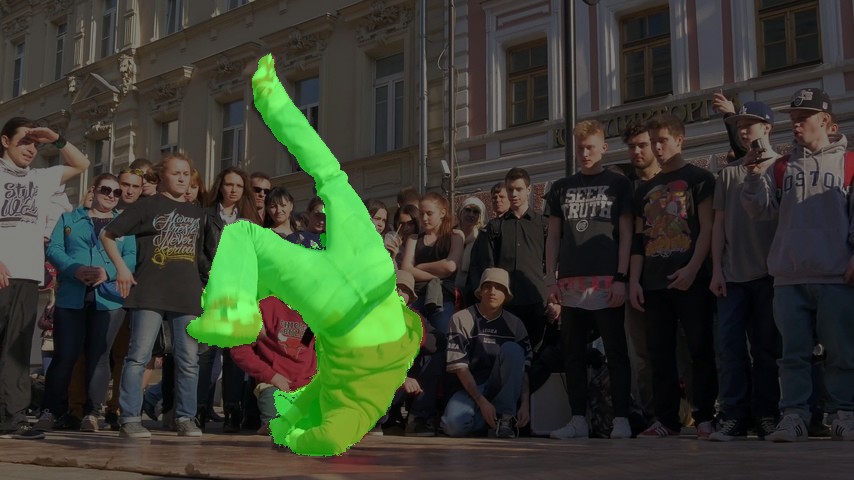}&
\includegraphics[width=.19\linewidth]{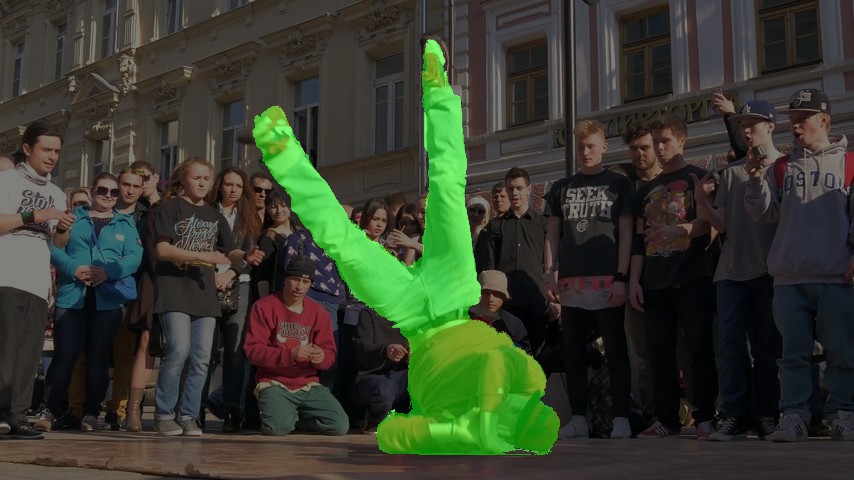}&
\includegraphics[width=.19\linewidth]{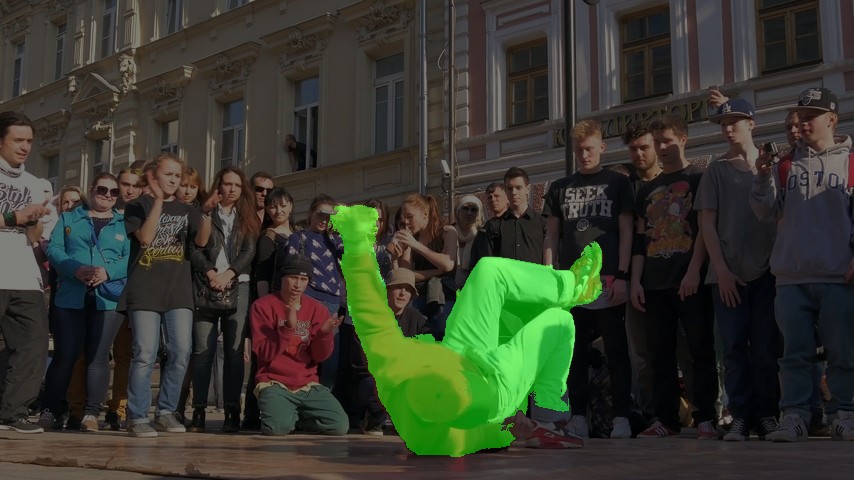}&
\includegraphics[width=.19\linewidth]{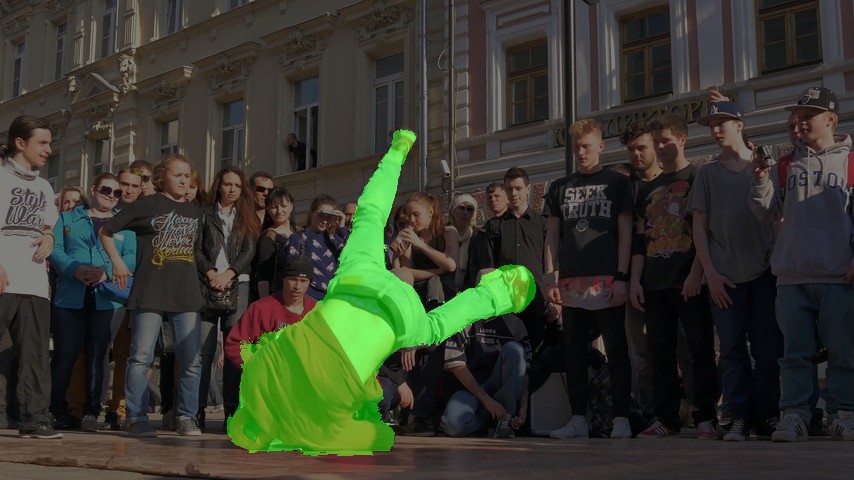}&
\includegraphics[width=.19\linewidth]{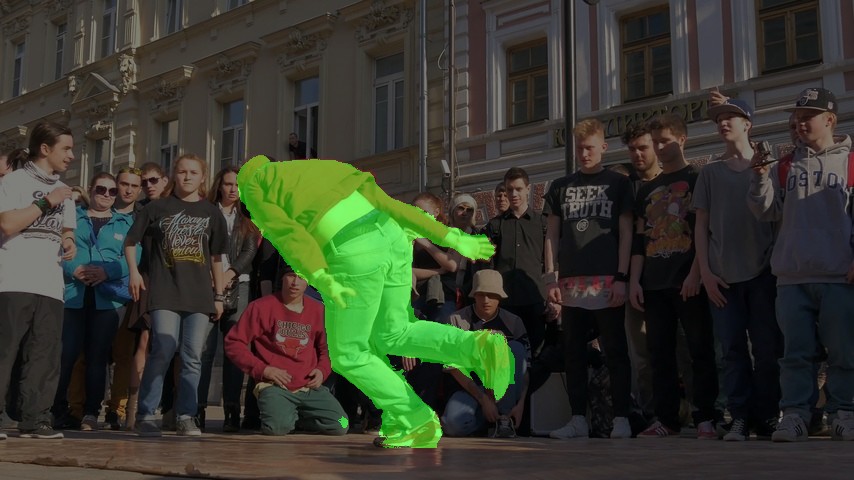}\\
\raisebox{.7\height}{\rotatebox[origin=t]{90}{Bmx-Trees}} &
\includegraphics[width=.19\linewidth]{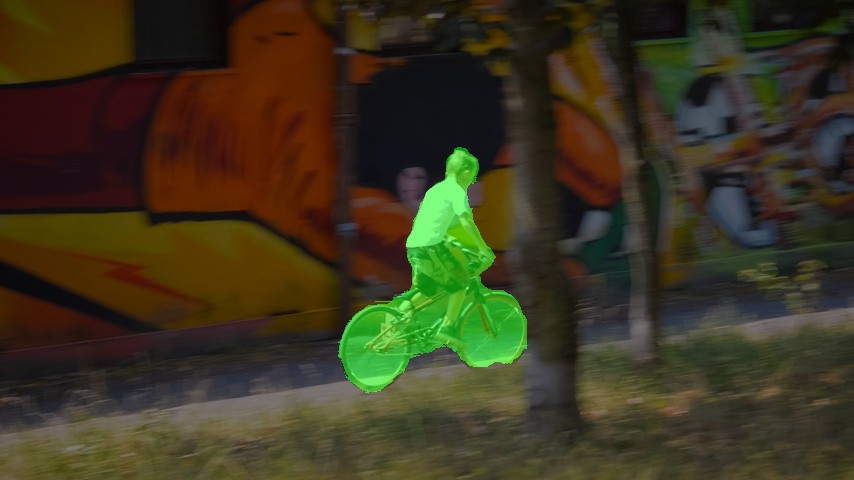}&
\includegraphics[width=.19\linewidth]{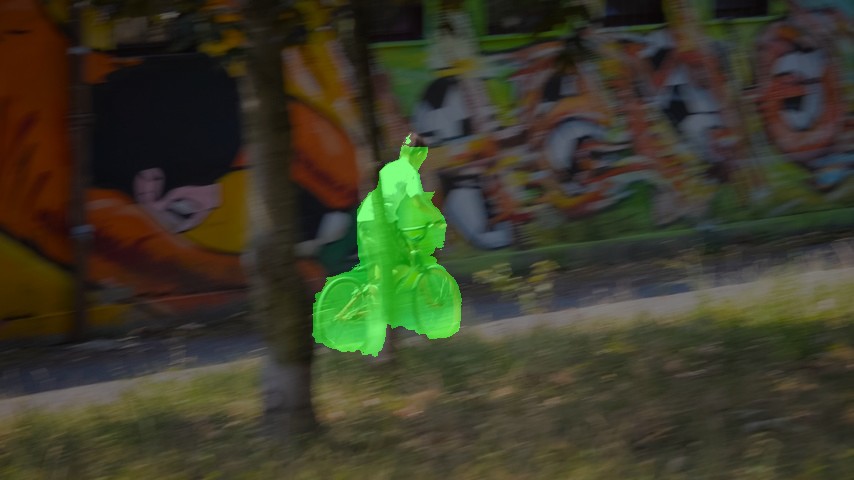}&
\includegraphics[width=.19\linewidth]{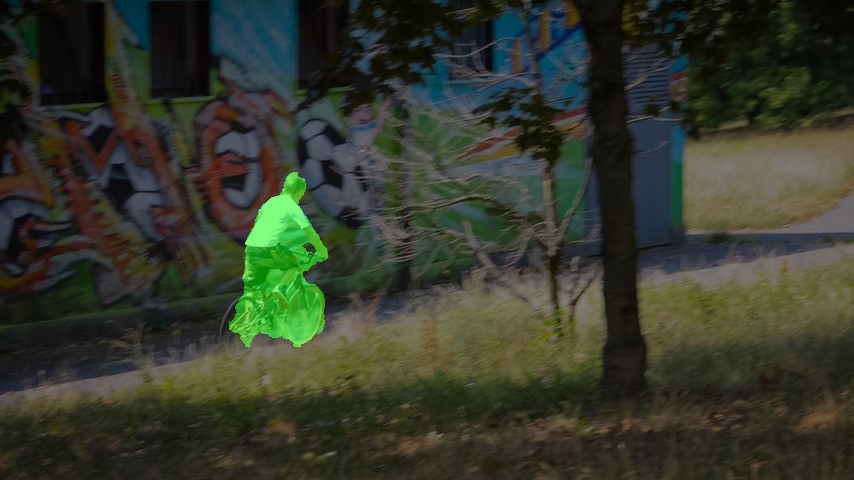}&
\includegraphics[width=.19\linewidth]{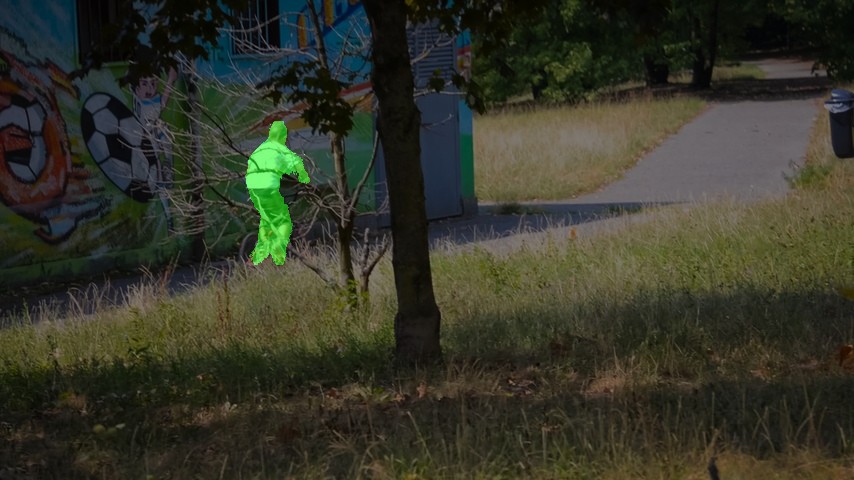}&
\includegraphics[width=.19\linewidth]{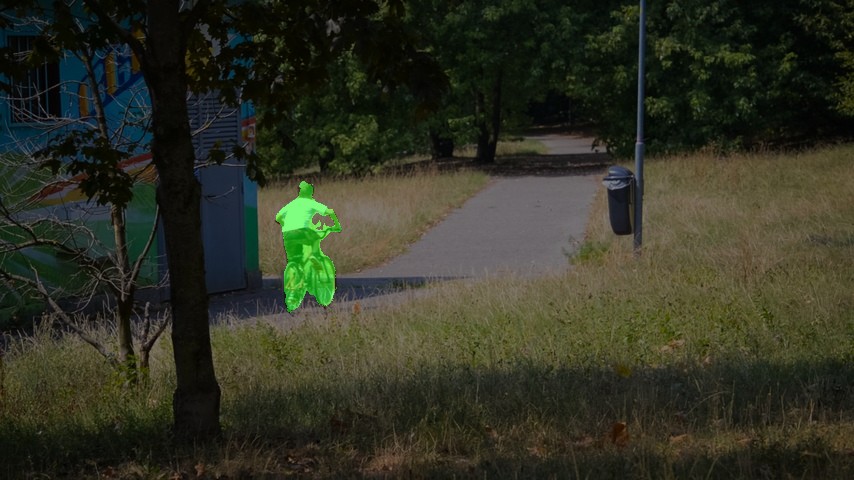}\\
\raisebox{.7\height}{\rotatebox[origin=t]{90}{Soapbox}} &
\includegraphics[width=.19\linewidth]{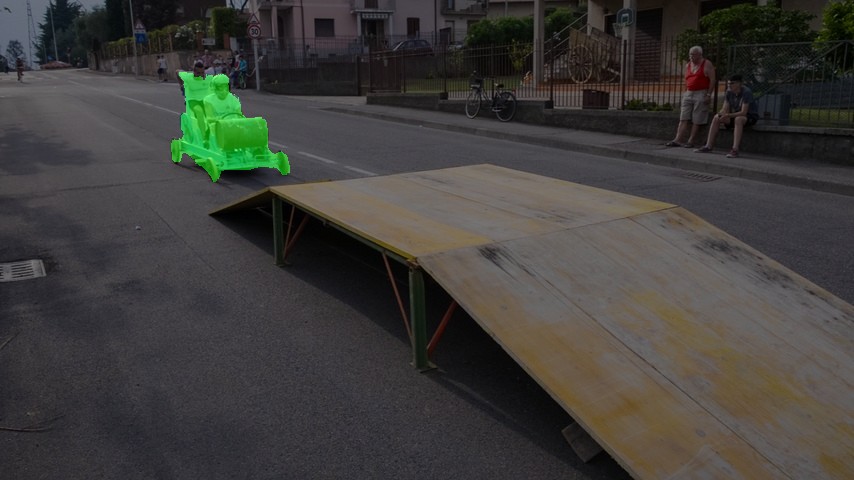}&
\includegraphics[width=.19\linewidth]{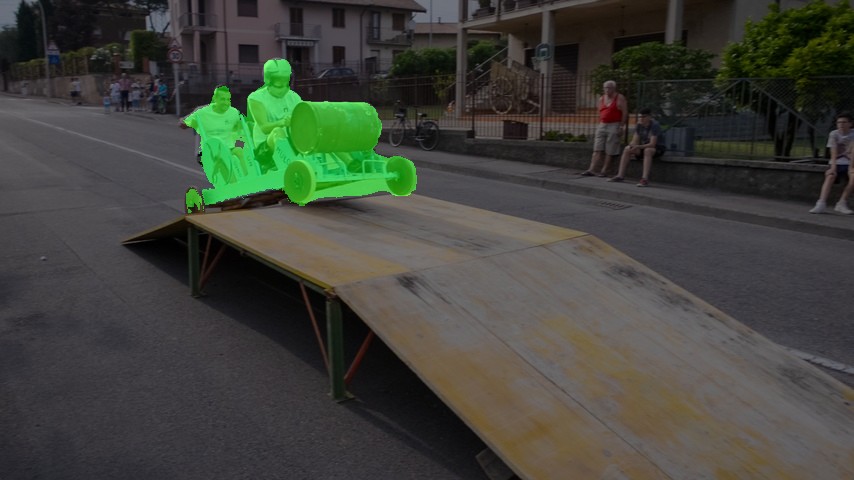}&
\includegraphics[width=.19\linewidth]{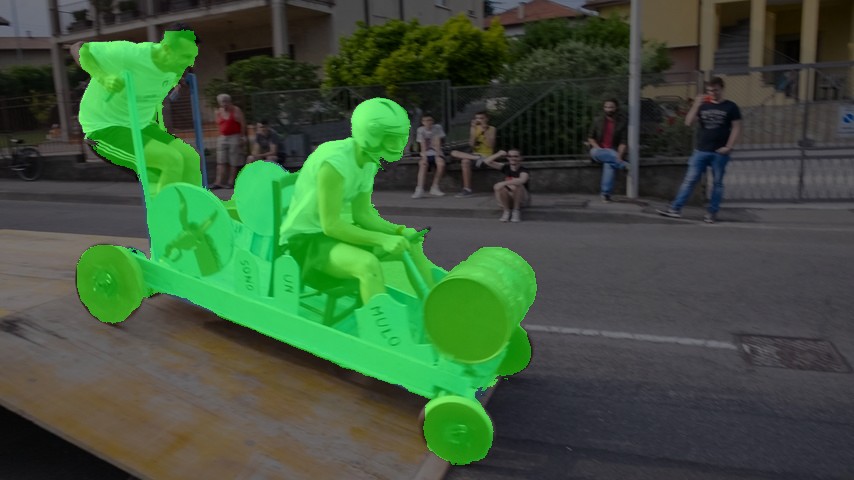}&
\includegraphics[width=.19\linewidth]{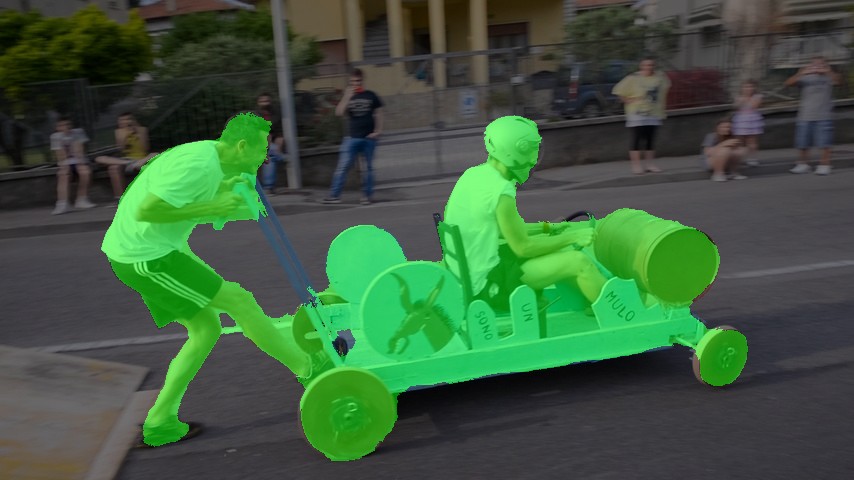}&
\includegraphics[width=.19\linewidth]{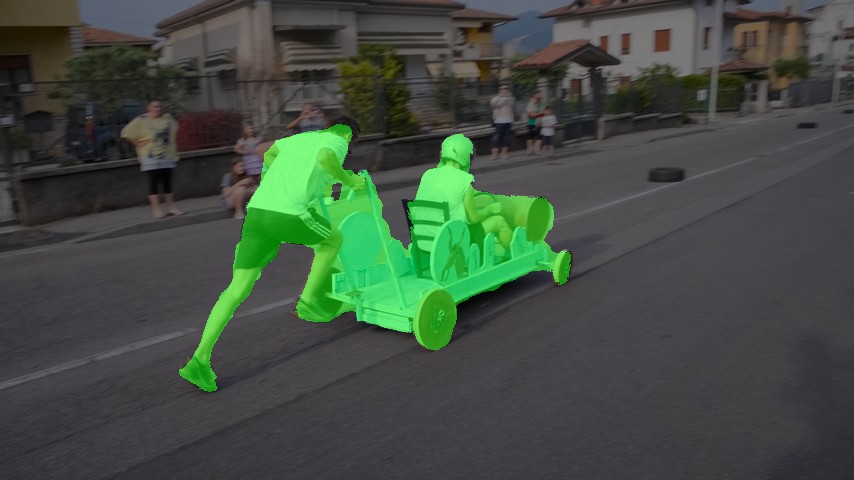}\\
\raisebox{.5\height}{\rotatebox[origin=t]{90}{Scooter-black}} &
\includegraphics[width=.19\linewidth]{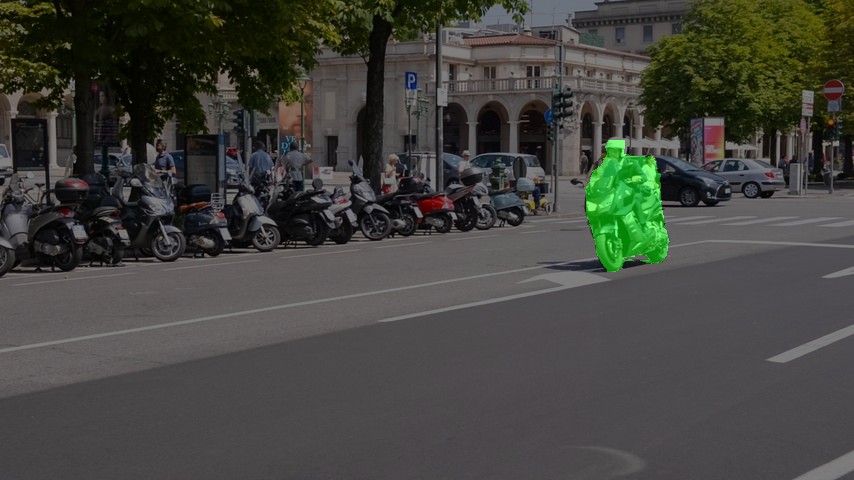}&
\includegraphics[width=.19\linewidth]{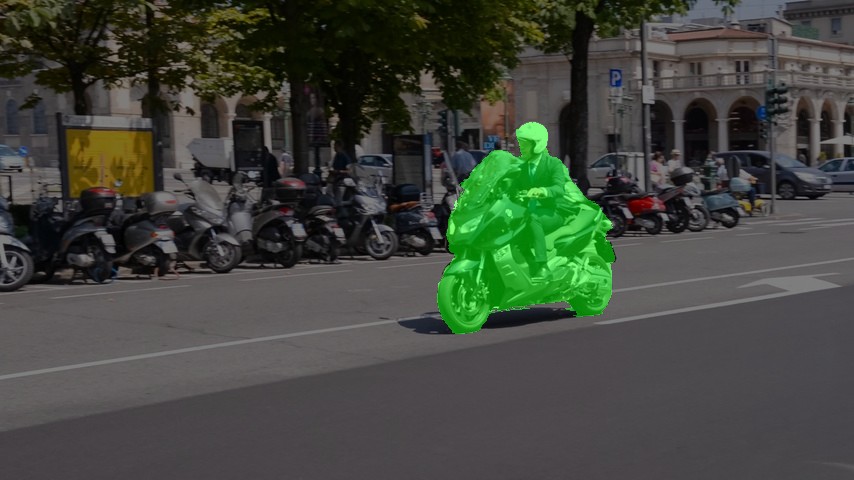}&
\includegraphics[width=.19\linewidth]{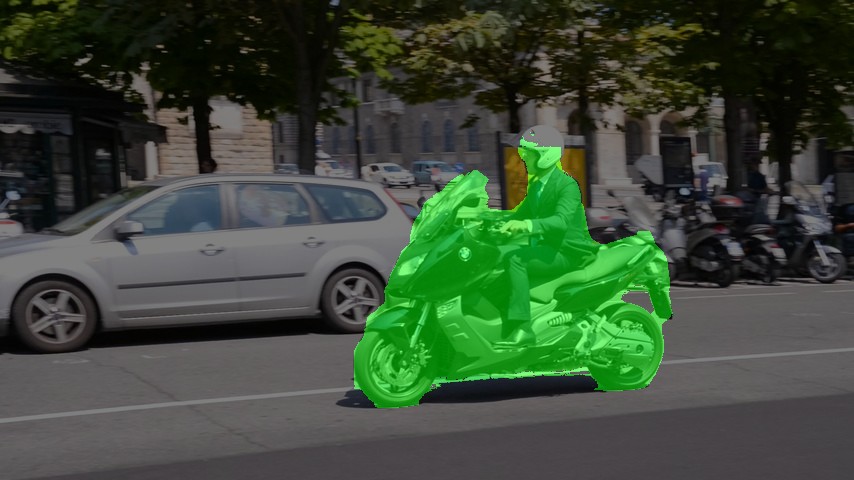}&
\includegraphics[width=.19\linewidth]{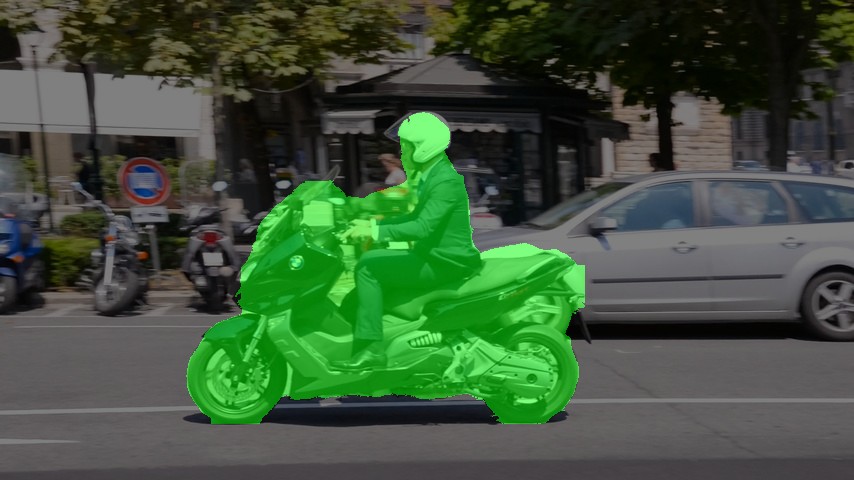}&
\includegraphics[width=.19\linewidth]{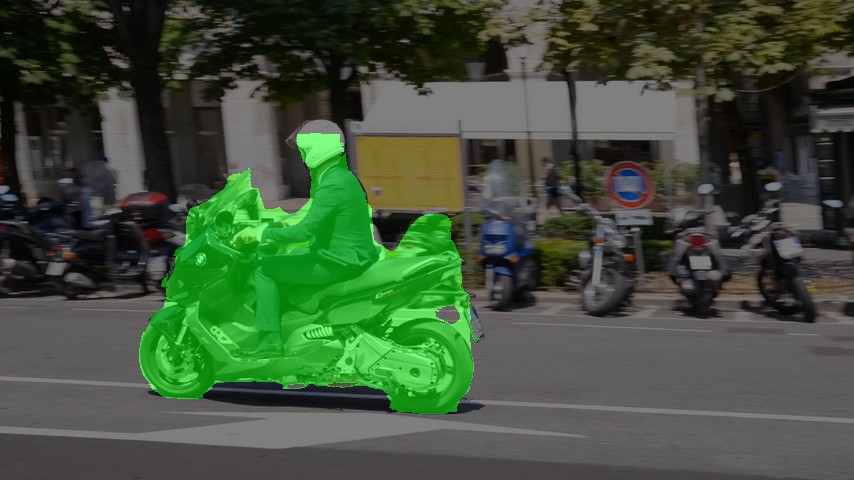}\\

\end{tabular}

\caption{Qualitative result of the proposed method in various sequences. The \textit{green} region is the segmentation result. 
% More detailed results can be found in the supplementary material.
}
\label{fig:our_results}
\end{figure*}

{\renewcommand{\arraystretch}{1.2} 
\begin{table*}[]
		\centering
\caption{Ablation study on training variants. Note that the evaluation metric is $\mathcal{J}\&\mathcal{F} $  mean, unlike Table~\ref{table:comparison}.}
% \begin{center}
\begin{tabular}{c|l|ccccccccc}
\hline

Aspect & Variant  & \multicolumn{8}{c}{Enable (\cmark) / Disable} \\
% & LSTM & Semantic & Time step & Instance & CRF & F-Measure \\
\hline 
\multirow{3}{*}{Stream}
                & Motion         &\cmark &        &      &\cmark &\cmark &\cmark&\cmark &\cmark &\cmark \\
                & Apprearance    &       &\cmark  &      &\cmark &\cmark &\cmark&\cmark &\cmark &\cmark \\
                & Instance       &       &        &\cmark&       &         &      &\cmark &\cmark &\cmark \\
\hline
Recurrent       & Cascaded       &       &        &      &       &\cmark &\cmark&\cmark &\cmark &\cmark \\
\hline
\multirow{2}{*}{Post-process}
                & CRF             &       &        &      &       &        &\cmark&         &\cmark &\cmark \\
                & Kalman         &       &        &      &       &       &       &         &       &\cmark \\

\Xhline{2\arrayrulewidth}

\multicolumn{2}{c|}{$\mathcal{J}\&\mathcal{F} $ Mean}            &69.30& 69.99& 77.82& 76.27& 77.06&78.76& 82.40& 83.87& 83.98 \\
% \multicolumn{2}{c|}{$\mathcal{J} $ Mean}                         &68.85& 69.91& 78.29& 76.15& 76.86&79.73& 82.78& 84.79& 84.91 \\        
% \multicolumn{2}{c|}{$\mathcal{F} $ Mean}                        &69.75& 70.07& 77.35& 76.39& 77.27&77.79& 82.03& 82.95& 83.05 \\                    
\hline
\end{tabular}
% \end{center}
\label{table:Ablation}
\end{table*}
}

\section{Experiments}

\subsection{Evaluation}

We evaluate our network on Densely Annotated VIdeo Segmentation (DAVIS)~\cite{davis2016} and Freiburg-Berkeley Motion Segmentation Dataset (FBMS-59)~\cite{14FBMS}. 

\subsubsection{DAVIS}
DAVIS has 50 videos divided into 30/20 training/validation sets. The metric of this dataset is region similarity $ \mathcal {J} $, contour accuracy $ \mathcal {F} $, and temporal stability $ \mathcal {T} $. Many studies compare performance by $ \mathcal {J} $ mean, and DAVIS 2017~\cite{davis2017} suggests the mean of the measure $ \mathcal {J} $ and $ \mathcal {F} $.
We verify the performance of the proposed structure by describing all possible metrics. Table~\ref{table:comparison} compares the results, including semi-supervised methods as well as unsupervised methods.
Our results show the performance gap about 4.5\%p in both $ \mathcal {J} $ and $ \mathcal {F} $ mean for the unsupervised method.
The unsupervised methods can be divided into the ones using end-to-end networks (LVO, PDB, Ours), and the others that segment embedded features in the graph (Embed, BNN, Mosal). LVO and PDB also proposed structures by combining deep CNN and recurrent network. So we suppose these two as the baselines for the proposed architecture. Compared to the performance of these methods (74.0, 75.6), the proposed method performs better by around 3\%p even when the instance information is excluded. This improvement demonstrates the effectiveness of multiscale stacked RNN and the fusion at the original resolution.
Our performance is lower than the performance of the state-of-the-art, but it is superior to most semi-supervised methods.
The qualitative results can be seen in Fig.~\ref{fig:our_results}.

\subsubsection{FBMS-59}
FBMS-59 consists of 29/30 sequences of train/test sets. Unlike DAVIS, this dataset is for multiple object detection. To verify the robustness of the proposed structure, the learned network for DAVIS is used for pixel-level segmentation without further learning. However, we use the pretrained instance proposal network for COCO instead of the newly learned network for DAVIS. This is because further learning to DAVIS is not trained to detect multiple proposals for different objects, resulting in less accurate proposal performance.
The results of unsupervised methods are summarized in Table~\ref{table:FBMS}. As a metric, the $ \mathcal {J} $ mean and F-score are used for comparison. The proposed method shows better performance in both metrics.
% The qualitative result can be seen in the bottom row of Fig.~\ref{fig:our_results}.

{\renewcommand{\arraystretch}{1.2} 
\begin{table*}[!t]

\caption{Overall segmentation results on DAVIS. The upper part of the table is the measurement in the FAVOS~\cite{Cheng_favos_2018} (they `\textit{parallelly use Titan X GPUs}') and the lower part is the measurement in our environment (single GPU - NVIDIA Titan Xp). We tested GitHub code of FAVOS for the comparison, but it took a much longer time ($\sim$ 6s) in our environment than the authors suggested. It is difficult to compare them directly, so we present each result separately. Our runtime includes the preprocessing time for optical flow computation. It is seen that our proposed method is one of the fastest segmentation algorithms.}
\begin{center}
\begin{tabular}{|l|c|c|c|c|c|c|c|}

\hline
Method&Initial mask&Pre-processing&Multi-GPU&Speed&$\mathcal{J}$ mean&$\mathcal{F}$ mean&$\mathcal{T}$ mean\\
\Xhline{2\arrayrulewidth}
OnAVOS~\cite{onavos}&\cmark & finetuning& & 13s& 0.861& 0.849& 0.190\\
Lucid~\cite{LucidDataDreaming_arXiv17}&\cmark & data,finetuning& & 40s& 0.848& 0.823& 0.158\\
FAVOS-ref~\cite{Cheng_favos_2018}&\cmark & no& \cmark& 1.8s& 0.824& 0.795& 0.263\\
OSVOS~\cite{osvos}&\cmark & finetuning& & 10s& 0.798& 0.806& 0.378\\
% MSK&\cmark & flow,finetuning& weak& 12s& 0.797& 0.754& 0.218\\
FAVOS-part~\cite{Cheng_favos_2018}&\cmark & no& \cmark& 0.60s& 0.779& 0.760& 0.229\\
ARP~\cite{ARP}& & data& & & 0.762& 0.706& 0.393\\
% SFL&\cmark & finetuning& weak& 7.9s& 0.761& 0.760& 0.189\\
LVO~\cite{Tokmakov17LVO}&\cmark & flow& & & 0.759& 0.721& 0.265\\
% CTN&\cmark & flow& weak& 29.95s& 0.735& 0.693& 0.220\\
FSEG~\cite{jain2017fusionseg} & & flow& & 7s& 0.707& 0.653& 0.328\\
\Xhline{2\arrayrulewidth}
\rowcolor{lightgray}
{FAVOS-part}~\cite{Cheng_favos_2018}&\cmark & no&  & 6s& 0.779& 0.760& 0.229\\
\rowcolor{lightgray}
\textbf{Ours-CRF}& & flow (included) &  & 0.7s& 0.849& 0.831& 0.187\\
\rowcolor{lightgray}
\textbf{Ours-noCRF}&  & flow (included)& & 0.3s& 0.828& 0.820&  - \\
\hline
\end{tabular}
\end{center}

\label{table:runtime}
\end{table*}
}
\subsection{Ablation studies}
\label{section:ablation}
\subsubsection{Recurrent architecture}
Although we can obtain motion information by optical flow alone, we confirm that using the network with longer temporal dependency improves the performance. As shown in Table~\ref{table:Ablation_RNN}, there are large performance gaps between the 2D convolution (intra-frame structure) and the rest of the inter-frame structures.
The architecture with the bidirectional cascade structure is also compared in the table. When we add a cascade, \ie~when the output of the forward network is transferred to the backward network, the performance is improved. This is because the structure with the cascade makes the recurrent stack of the entire motion stream deeper (in our case, 10 RNN stacks) and makes the size of the timescale more variable.

\subsubsection{Stream ablation}
To analyze the effect of each part of the proposed structure, we experiment with various ablations such as omitting each part or putting it alone, and summarize the results in Table~\ref{table:Ablation}. First, we can see that the motion/appearance stream of pixel-level segmentation shows similar performance when each is trained separately but shows improvement when the two are combined. It can be assumed that each stream has complementary information.
The 4-th result of Table~\ref{table:Ablation} is the performance of pixel-level segmentation without instance-aware segmentation and other post-processing.
Without the iterative operation or the handcrafted structure, the encoder-decoder stream fusion performs favorably against existing state-of-the-art algorithms.
The performance of instance proposal is similar to the pixel-level result. This shows that instance-level information is important in video object segmentation. The instance-aware segmentation significantly improves the performance by effectively integrating different levels of information (shown in the rightmost column).

\subsubsection{Instance-aware segmentation ablation}
\begin{figure}
\begin{center}

\setlength\tabcolsep{0.5 pt}
\begin{tabular}{lccccc}
\includegraphics[width=.49\linewidth]{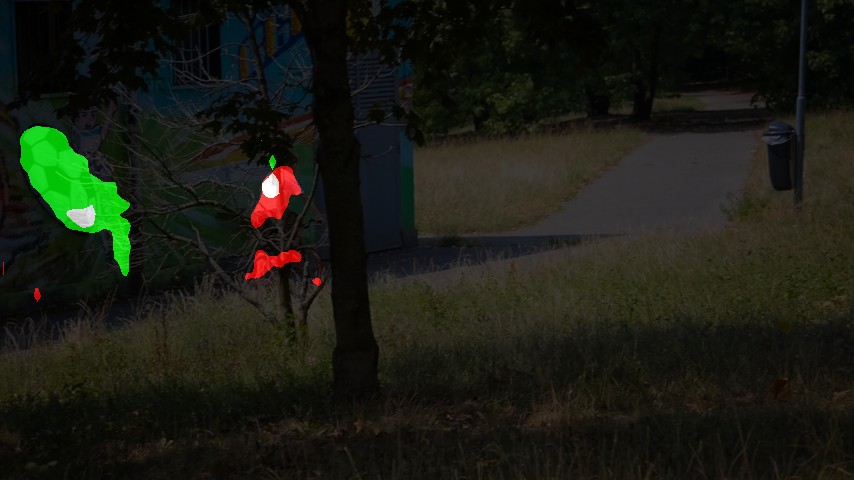}&
\includegraphics[width=.49\linewidth]{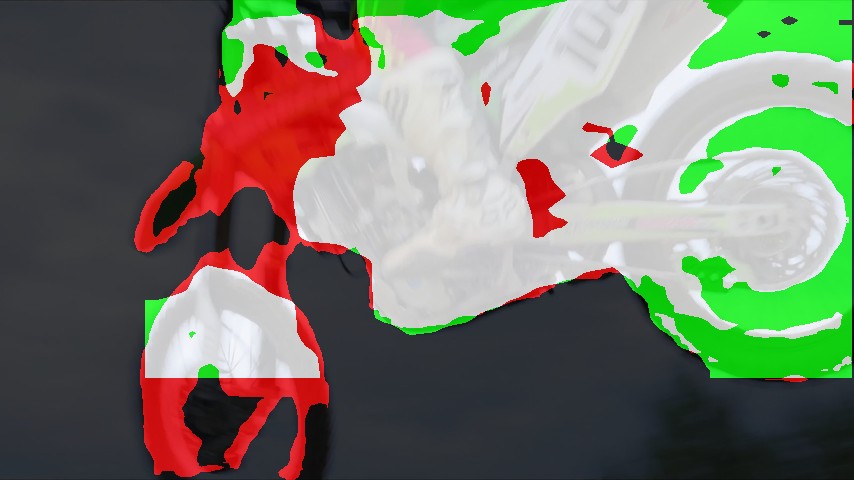}\\
\end{tabular}
\end{center}
\vspace{-10pt}
   \caption{Failure case of the instance-aware segmentation: \textit{red}, \textit{green}, \textit{white} regions are results of the pixel-level segmentation, the instance-level segmentation and the intersection of the two.}
\label{fig:instance_failure}
\end{figure}

The easiest way to use object information is to select a single object with the highest probability. Since this method loses pixel-level information, it exhibits similar performance (82.32 vs. 82.40) before post-processing, but there is little to gain after post-processing (82.33 vs. 83.98). This is an average of 1.6\%p difference, but it causes a large error in some frames when object detection fails or partially missed, as Fig.~\ref{fig:instance_failure}.
Our instance-aware segmentation complements two results, and thus improves the performance.

\subsection{Runtime analysis}
The proposed structure spends the most time in optical flow and instance proposal calculation. For the accuracy of the result, the optical flow operation uses 480p. This operation requires about 120 ms per frame in the NVIDIA Titan Xp. The instance proposal takes the same resolution and takes about 130 ms per frame. However, the subsequent procedures use the preprocessed image so that the inference is fast.
In the same GPU environment, the pixel-level segmentation takes 580 ms for 20 frame mini-batch, and it takes less than 40 ms to calculate a frame with the largest I/O overhead. The instance-aware segmentation takes 25 ms including Kalman filter. The total elapsed time of each step is about 300 ms.
Although the direct comparison is not possible due to differences in the GPU environment, our proposed method is one of the fastest segmentation algorithms compared to the speed of other methods described in~\cite{Cheng_favos_2018}. Comparisons of runtime are given in Table~\ref{table:runtime}.

\section{Conclusion}

In this paper, we have proposed an instance-aware video segmentation method that does not require any annotation. The proposed multiscale recurrent network learns motion information intensively by acquiring various timescales with a deep stack of recurrent networks. The encoder-decoder stream fusion combines the information in each stream without loss of resolution and interference between streams. Finally, instance-aware segmentation ensures the pixel-level result to be integrated with the instance proposal. The proposed method outperforms the existing unsupervised methods and shows comparable performance to the state-of-the-art semi-supervised methods on DAVIS and FBMS-59.
Also, our method is one of the fastest video object segmentation algorithms.
Our proposed method has achieved good performance in FBMS-59 with many objects.
This reveals the possibility of our method in automatic multiple object segmentation, and this is our future work. We will make our code publicly available.

\section*{Acknowledgment}

The authors would like to thank...

\ifCLASSOPTIONcaptionsoff
  \newpage
\fi

{\small
\bibliographystyle{IEEEtran}
\bibliography{egbib}
}

\end{document}